\documentclass{article}

\usepackage[nonatbib, preprint]{neurips_2023}

\usepackage[utf8]{inputenc} 
\usepackage[T1]{fontenc}    
\usepackage{hyperref}       
\usepackage{url}            
\usepackage{booktabs}       
\usepackage{amsfonts}       
\usepackage{nicefrac}       
\usepackage{microtype}      
\usepackage{xcolor}         
\usepackage{multirow}
\usepackage{graphicx}
\usepackage{authblk}
\usepackage{makecell}
\usepackage{subcaption}
\usepackage{pifont}
\usepackage{comment}
\usepackage{algorithm}
\usepackage{algorithmic}
\usepackage{amsmath}

\title{SEED-Bench: Benchmarking Multimodal LLMs with Generative Comprehension}

\begin{document}

\author{
\textbf{Bohao Li$^{1\star}$  \ Rui Wang$^{1\star}$ \ Guangzhi Wang$^{2\star}$  \ Yuying Ge$^{1\dagger}$  \ Yixiao Ge$^{1,2\dagger}$ \ Ying Shan$^{1,2}$} 

$^{1}$Tencent AI Lab \qquad $^{2}$ARC Lab, Tencent PCG

\url{https://github.com/AILab-CVC/SEED-Bench}
}
 \renewcommand{\thefootnote}{\fnsymbol{footnote}}
 		\footnotetext[1]{Equal Contribution.} 
   \footnotetext[2]{Correspondence to \texttt{yuyingge@tencent.com} and \texttt{yixiaoge@tencent.com}.}
   
\maketitle

\begin{abstract}
    Based on powerful Large Language Models (LLMs), recent generative Multimodal Large Language Models (MLLMs) have gained prominence as a pivotal research area, exhibiting remarkable capability for both comprehension and generation. In this work, we address the evaluation of generative comprehension in MLLMs as a preliminary step towards a comprehensive assessment of generative models, by introducing a benchmark named SEED-Bench. SEED-Bench consists of 19K multiple choice questions with accurate human annotations ($\times$6 larger than existing benchmarks), which spans 12 evaluation dimensions including the comprehension of both the image and video modality. We develop an advanced pipeline for generating multiple-choice questions that target specific evaluation dimensions, integrating both automatic filtering and manual verification processes. Multiple-choice questions with groundtruth options derived from human annotation enables an objective and efficient assessment of model performance, eliminating the need for human or GPT intervention during evaluation. We further evaluate the performance of 18 models across all 12 dimensions, covering both the spatial and temporal understanding. By revealing the limitations of existing MLLMs through evaluation results, we aim for SEED-Bench to provide insights for motivating future research. We will launch and consistently maintain a leaderboard to provide a platform for the community to assess and investigate model capability.
\end{abstract}

\section{Introduction}

\begin{figure}
    \centering
    \begin{subfigure}{0.62\textwidth}
        \centering
        \includegraphics[width=\linewidth]{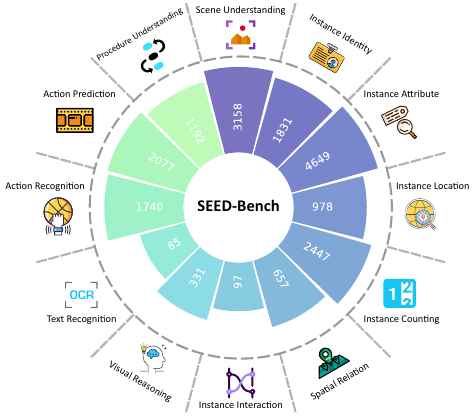}
    \end{subfigure}
    \hfill
    \begin{subfigure}{0.33\textwidth}
        \centering
        \includegraphics[width=\linewidth]{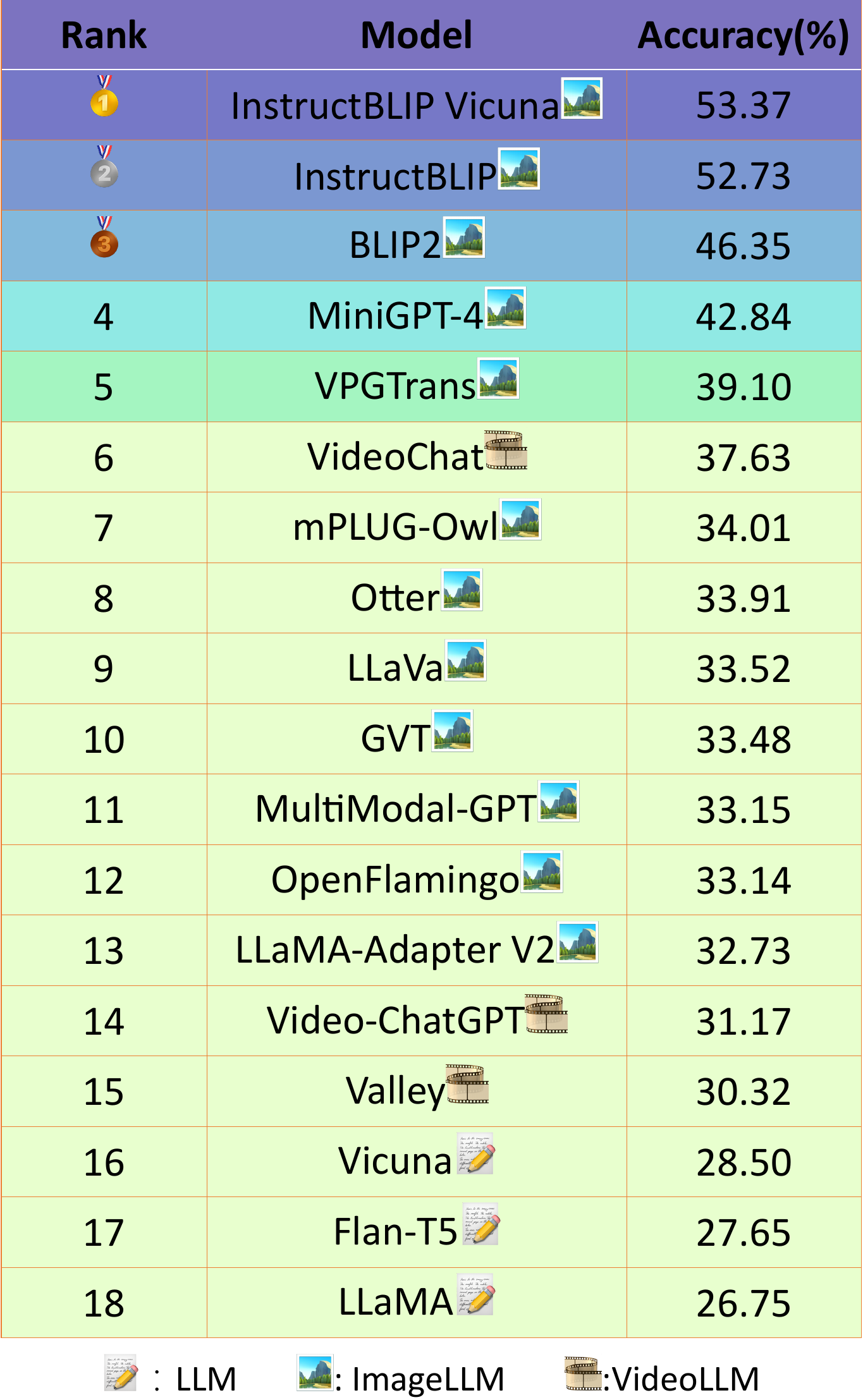}
    \end{subfigure}
    \vspace{5pt}
    \caption{Left: Overview of 12 evaluation dimensions in SEED-Bench including both the spatial and temporal understanding, where the number in the bar denotes the number of human-annotated multiple-choice questions in each dimension. Right: the overall leaderboard displaying the averaged accuracy of 18 models across 12 evaluation dimensions.}
    \label{fig:seed_bench_overview}
\end{figure}

In recent years, Large Language Models (LLMs)~\cite{chung2022scaling_flant5, openai2023gpt4, ChatGPT, vicuna, touvron2023llama} have exhibited remarkable capabilities to understand, reason, and generate texts across a variety of open-ended tasks. Leveraging the strong generality of LLMs, generative Multimodal Large Language Models (MLLMs)~\cite{li2023blip2, zhu2023minigpt4, liu2023visual_llava, ye2023mplugowl, dai2023instructblip, li2023otter, gong2023multimodalgpt, su2023pandagpt, peng2023kosmos, li2023videochat, maaz2023videochatgpt, luo2023valley, ge2023planting, sun2023generative,yu2023scaling, koh2023gill} have demonstrate enhanced abilities for 
multimodal comprehension and generation. However, current MLLMs mainly evaluate their performance with a limited number of qualitative examples, or by employing previous benchmarks that are not tailored for evaluating MLLMs with open-form output. For example, in VQAv2~\cite{goyal2017making}, an answer is considered correct only if the model's output exactly matches the groundtruth answer, which typically consists of just one or two words. The lack of a comprehensive and objective benchmark to evaluate MLLMs poses a significant challenge for comparing and investigating the performance of various models.

Concurrent works~\cite{fu2023mme,yin2023lamm,xu2023lvlm,liu2023mmbench} have made efforts to develop benchmarks for specifically evaluating MLLMs as shown in Table~\ref{tab:benchmark_compare}. For example, LVLM-eHub~\cite{xu2023lvlm} and LAMM~\cite{yin2023lamm} utilize exiting public datasets across various computer vision tasks as evaluation samples, and employ human annotators or GPT to assess the quality, relevance, and usefulness of model's predictions. However, the involvement of human and GPT during evaluation not only compromises efficiency, but also leads to increased subjectivity and reduced accuracy of the assessment. MME~\cite{fu2023mme} and MMBench~\cite{liu2023mmbench} further advance objective evaluation of MLLMs by constructing True/False Questions or Multiple-Choice Questions, which cover a variety of ability dimensions. Restricting the model's output to True/False or A/B/C/D options facilitates the convenient computation of accuracy, which serves as an objective metric for evaluation. However, the relatively small scale of these benchmarks (fewer than 3K samples) introduces instability in the evaluation statistics.

\begin{figure}
    \includegraphics[width=1.0\textwidth]{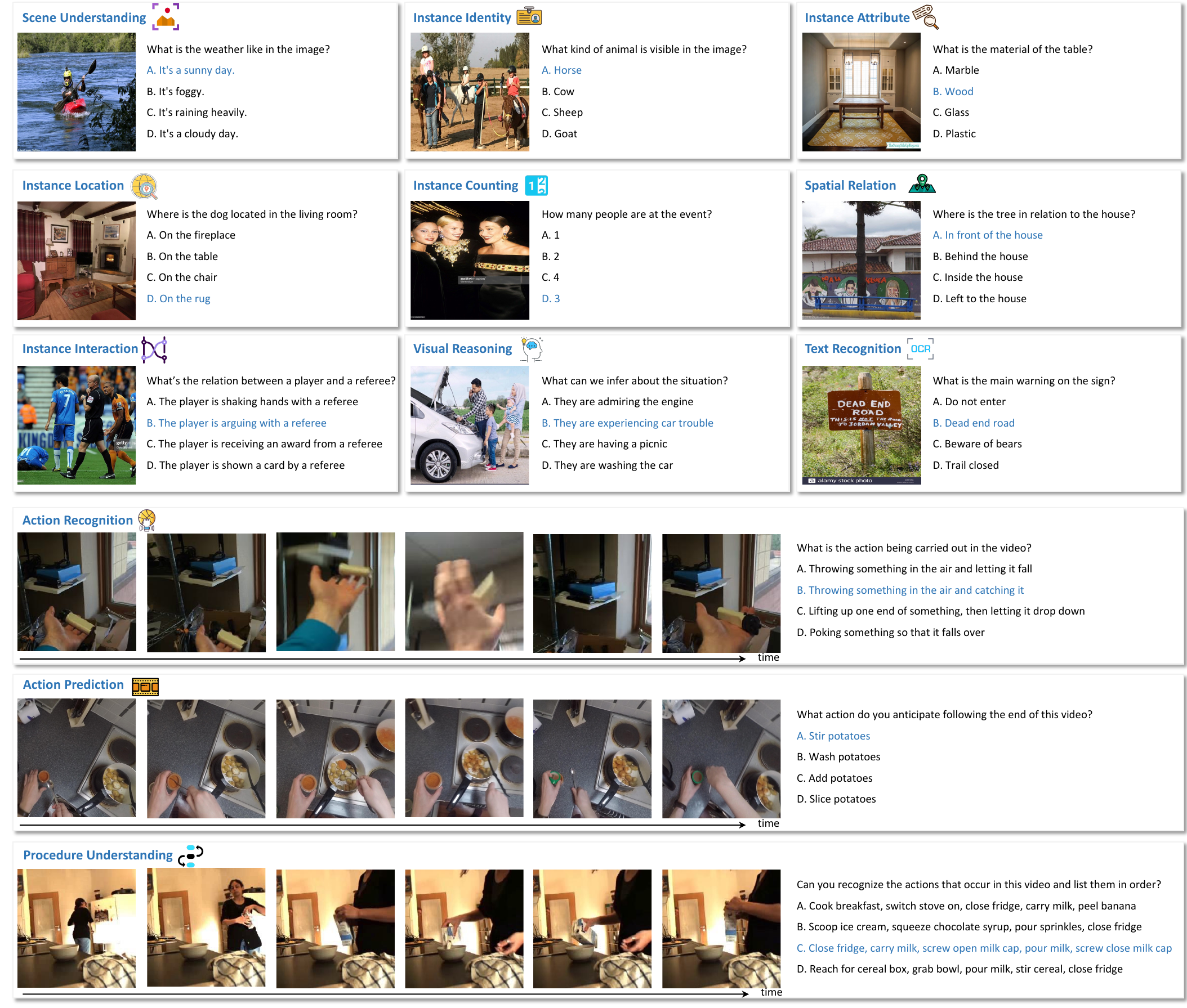}
    \caption{Data samples of SEED-Bench, which covers 12 evaluation dimensions including both the spatial and temporal understanding. Each evaluation dimension contains multiple-choice questions with groundtruth options derived from human annotation.}
    \label{fig:example}
        \vspace{-12pt}
\end{figure}

In this work, we focus on evaluating the generative comprehension capability of MLLMs as a preliminary step towards a comprehensive assessment of generative
models, by introducing a benchmark named SEED-Bench\footnote{In pursuit of Artificial General Intelligence (AGI), LLMs have witnessed substantial progress. We have made a bold assumption that the premise for the emergence of multimodal capabilities is to unify both comprehension and generation within an autoregressive generative model, where SEED~\cite{ge2023planting} takes a modest step. Besides the exploration of models, it is essential to have appropriate evaluations that motivate research directions. Therefore, we concurrently propose SEED-Bench to evaluate the comprehension ability of generative models.}. SEED-Bench spans 12 evaluation dimensions across both image and video modalities as shown in Fig.~\ref{fig:seed_bench_overview}. SEED-Bench consists of 19K multiple choice questions with groundtruth answers derived from human annotation ($\times$9 larger than MME and $\times$6 larger than MMBench) as shown in Fig.~\ref{fig:example}. We design a sophisticated pipeline for the generation of multiple-choice questions that are tailored to evaluate specific dimensions. We further incorporate automated filtering mechanism and manual verification process to ensure the quality of questions and the accuracy of groundtruth answers.   

Specifically, for images, we utilize various foundation models to extract their visual information including image-level captions~\cite{li2023blip2, huang2023tag2text}, instance-level descriptions~\cite{wu2022grit,kirillov2023sam,zhang2021vinvl} and textual  elements~\cite{paddleocr}. For videos, we leverage the original human annotations to provide visual information. We then feed the visual information to ChatGPT/GPT-4 with specially designed prompts corresponding to specific evaluation dimension. ChatGPT/GPT-4 subsequently generates questions as well as four candidate options with one groundtruth answer. We further filter out questions that can be answered without the visual input through utilizing multiple LLMs. Finally, we employ human annotators to choose the correct option of each multiple-choice question and classify each question into one evaluation dimension, resulting in a clean and high-quality benchmark containing 19K multiple-choice questions. Our pipeline supports the scalability of evaluation data across multiple domains, and we will continue to expand the benchmark with more evaluation dimensions.

\begin{table}[]
    \centering
    \caption{Comparisons between existing benchmarks for Multimodal LLMs. ``H/G Evaluation'' denotes whether human or GPT is used for evaluation.}\label{tab:benchmark_compare}
    \vspace{3pt}
    {\small
    \resizebox{\textwidth}{!}{
    \begin{tabular}{cccccccc}
         \toprule
             Benchmark & Visual Modality & Customized Question & \#Answer Annotation& Answer Type & H/G Evaluation & \#Models \\
         \midrule
         MME~\cite{fu2023mme} & Image   & \ding{51} & 2194 &Y/N & N/A & 10\\
         LAMM~\cite{yin2023lamm} & Image \& Point cloud  & \ding{55} & - & free-form & GPT & 4\\
         LVLM-eHub~\cite{xu2023lvlm} & Image   & \ding{55} & - & free-form & Human &8\\
         MMBench~\cite{liu2023mmbench} & Image   &\ding{51} &2974 & free-form& GPT &14\\
         Ours & Image \& Video  & \ding{51} & 19242 & A/B/C/D&  N/A & 18\\
         \bottomrule
    \vspace{-20pt}
    \end{tabular}
    }
   }
\end{table}

Based on SEED-Bench, we comprehensively evaluate 18 models including LLMs, ImageLLMs and VideoLLMs across all 12 dimensions as shown in Fig.~\ref{fig:seed_bench_overview}. Different from MMBench~\cite{liu2023mmbench} that employs ChatGPT to match a model’s prediction to one of the choices in a multiple-choice question (achieves only 87.0\% alignment rate), we follow GPT-3~\cite{brown2020gpt3} to calculate log-likelihood for each candidate option and select the one with the highest value as the final prediction, without relying on the instruction-following capabilities of models to output ``A'' or ``B'' or ``C'' or ``D''. By analyzing the results across 12 dimensions, we conduct a comprehensive comparison of existing multimodal models in both spatial and temporal understanding capabilities. We observe that the majority of MLLMs still exhibit limited performance across all 12 evaluation dimensions, and surprisingly find that VideoLLMs fail to achieve competitive performance on temporal understanding compared with ImageLLMs. Through the evaluation results, we aim for SEED-Bench to provide insights for motivating future exploration of a more advanced MLLM. We will launch an evaluation platform and consistently maintain a leaderboard for assessing and comparing model performance.

\section{Related Work}\label{sec:related_work}
\noindent\textbf{Multimodal Large Language Models.}
With the impressive success of Large language models (LLM)~\cite{chung2022scaling_flant5,touvron2023llama,vicuna}, recent studies work on generative Multimodal Large Language Models (MLLMs)~\cite{li2023blip2, zhu2023minigpt4, liu2023visual_llava, ye2023mplugowl, dai2023instructblip, li2023otter, gong2023multimodalgpt, su2023pandagpt, peng2023kosmos, ge2023planting, sun2023generative,yu2023scaling, koh2023gill} to improve multimodal comprehension and generation through utilizing the strong generality of LLMs. Some work~\cite{li2023videochat,maaz2023videochatgpt,luo2023valley} further considers video inputs and leverage the vast capabilities of LLMs for video understanding tasks. In SEED-Bench, we provide a comprehensive quantitative evaluations of these models to thoroughly assess and compare their performance in generative comprehension.

\noindent\textbf{Benchmarks for Multimodal Large Language Models.}
With the rapid development of Multimodal Large Language Models (MLLMs), some concurrent works~\cite{fu2023mme,yin2023lamm,xu2023lvlm,liu2023mmbench} propose various benchmarks for evaluating MLLMs. For example, GVT~\cite{wang2023gvt} constructs a benchmark by aggregating two semantic-level understanding tasks (VQA and Image Captioning) and two fine-grained tasks (Object Counting and Multi-class Identification). But its evaluation is constrained to limited aspects of visual understanding. LVLM-eHub~\cite{xu2023lvlm} combines multiple existing computer vision benchmarks and develops an online platform, where two models are prompted to answer a question related to an image and human annotators are employed to compare the predictions of models. The involvement of human annotators during evaluation not only introduces bias but also incurs significant costs. LAMM~\cite{yin2023lamm} evaluates image and point cloud tasks by using entity extraction to obtain key answers from open-form predictions and utilizing GPT to evaluate the answers’ relevance and accuracy to the groundtruth. The reliance on entity extraction and GPT metric can impact the accuracy and reliability of the evaluation. MME~\cite{fu2023mme} and MMBench~\cite{liu2023mmbench} aim to enhance the objective evaluation of MLLMs by constructing 2914 True/False Questions and 2974 Multiple Choice Questions across a variety of ability dimensions respectively. Considering the relatively small scale of these benchmarks, their evaluation results may exhibit instability. In this work, we introduce SEED-Bench to provide objective and comprehension evaluation of MLLMs, which contains 19K multiple-choice questions covering 12 evaluation dimensions including both spatial and temporal understanding.

\section{SEED-Bench}
Our benchmark contains 19K multiple-choice questions with accurate human annotations spanning 12 evaluation dimensions including both the spatial and temporal understanding. In this section, we first present the evaluation dimensions of SEED-Bench in Sec.~\ref{sec:pipeline_multi_task}. We introduce the data source in Sec.~\ref{sec:source} and our pipeline for constructing multiple-choice questions in Sec.~\ref{sec:mcq}. We finally describe the evaluation strategy for MLLMs to answer multiple-choice questions in Sec.~\ref{sec:strategy}.

\begin{table}[]
    \centering
    \caption{Evaluation dimensions of SEED-Bench including both the spatial and temporal understanding. We omit the image in the sample questions.}\label{tab:multilevel}
    \vspace{6pt}
    {\small
    \resizebox{\textwidth}{!}{
    \begin{tabular}{cll}
         \toprule
         & Evaluation Dimensions & Sample Questions\\ 
         \midrule
         \multirow{25}{*}{Spatial Understanding} & 1. Scene Understanding & \makecell[l]{What is the weather like in the image? \\
          A. It's a sunny day 
          B. It's foggy  \\
          C. It's raining heavily 
          D. It's a cloudy day}
         \\
         \cmidrule(lr){2-3} 
         & 2. Instance Identity & \makecell[l]{What kind of animal is visible in the image? \\
         A. Horse 
         B. Cow 
         C. Sheep 
         D. Goat}\\
         \cmidrule(lr){2-3}
         & 3. Instance Attribute & \makecell[l]{What is the material of the table? \\
         A. Marble 
         B. Wood 
         C. Glass 
         D. Plastic} \\
         \cmidrule(lr){2-3} 
         & 4. Instance Location & \makecell[l]{Where is the dog located in the living room? \\
         A. On the fireplace 
         B. On the table 
         C. On the chair 
         D. On the rug }\\ 
         \cmidrule(lr){2-3} 
         &5. Instance Counting & \makecell[l]{How many people are there in the image? \\
         A. 1 
         B. 2 
         C. 4 
         D. 3}\\
        \cmidrule(lr){2-3}
         &6. Spatial Relation  & \makecell[l]{What is the tree in relateion to the house? \\
            A. In front of the house 
            B. Behind the house \\
            C. Inside the house 
            D. Left to the house 
         }\\
         \cmidrule(lr){2-3}
         &7. Instance Interaction            & \makecell[l]{
         What is the relation between a player and a referee? \\
            A. The player is shaking hands with a referee \\
            B. The player is arguing with a referee \\
            C. The player is receiving an award from a referee \\
            D. The player is shown a card by a referee 
         } \\
         \cmidrule(lr){2-3} 
         &8. Visual Reasoning & \makecell[l]{what can we infer about the situation? \\
                    A. They are admiring the engine 
                    B. They are experiencing car trouble \\
                    C. They are having a picnic 
                    D. They are washing the car }\\
        \cmidrule(lr){2-3} 
        &9. Text Recognition &  \makecell[l]{
         What is the main warning on the sign? \\
            A. Do not enter 
            B. Dead end road \\
            C. Beware of bears 
            D. Trail closed}\\
        \midrule
        \multirow{10}{*}{Temporal Understanding} &10. Action Recognition & \makecell[l]{What is the action being carried out in the video? \\
         A. Throwing something in the air and letting it fall \\
         B. Throwing something in the air and catching it \\
         C. Lifting up one end of something, then letting it drop down \\
         D. Poking something so that it falls over}\\
        \cmidrule(lr){2-3}
        &11. Action Prediction  & \makecell[l]{What action do you anticipate following the end of this video? \\
            A. Stir potatoes
            B. Wash potatoes 
            C. Add potatoes 
            D. Slice potatoes
         }\\
         \cmidrule(lr){2-3}
         &12. Procedure Understanding            & \makecell[l]{
         Can you recognize the actions in this video and list them in order? \\
            A. Cook breakfast, switch stove on, close fridge, carry milk, peel banana \\
            B. Scoop ice cream, squeeze chocolate syrup, pour sprinkles, close fridge \\
            C. Close fridge, carry milk, screw open milk cap, pour milk, screw close milk cap \\
            D. Reach for cereal box, grab bowl, pour milk, stir cereal, close fridge \\
         } \\
         \bottomrule
    \end{tabular}
    }
   }
\end{table}

\subsection{Evaluation Dimensions}
\label{sec:pipeline_multi_task}

In order to comprehensively assess the visual understanding capability of MLLMs, SEED-Bench incorporates 12 evaluation dimensions including both the spatial and temporal comprehension as shown in Table~\ref{tab:multilevel}. 

\noindent\textbf{Spatial Understanding.} For the evaluation of spatial comprehension, we consider 9 dimensions covering image-level and instance-level perception and reasoning.
\begin{itemize}
    \item Scene Understanding. This dimension focuses on the global information in the image. Questions can be answered through a holistic understanding of the image.
    \item Instance Identity. This dimension involves the identification of a certain instance in the image, including the existence or category of a certain object in the image. It evaluates a model's object recognition capability.  
    \item Instance Attributes. This dimension is related to the attributes of an instance, such as color, shape or material. It assesses a model's understanding of an object's visual appearance. 
    \item Instance Location. This dimension concerns the absolute position of one specified instance. It requires a model to correctly localize the object referred to in the question.
    \item Instances Counting. This dimension requires the model to count the number of a specific object in the image. This requires the model to understand all objects, and successfully count the referred object's instances.
    \item Spatial Relation. This dimension asks an model to ground the two mentioned objects, and recognize their relative spatial relation within the image.
    \item Instance Interaction. This dimension requires the model to recognize the state relation or interaction relations between two humans or objects.
    \item Visual Reasoning. This dimension evaluates if a model is able to reason based on the visual information. This requires the model to fully understand the image and utilize its commonsense knowledge to correctly answer the questions.
    \item Text Understanding. For this dimension, the model should answer question about the textual elements in the image.
    \end{itemize}
\noindent\textbf{Temporal Understanding.} For the evaluation of temporal comprehension, we consider 3 dimensions focusing on the recognition, prediction and procedure understanding of actions.
\begin{itemize}
    \item Action Recognition. In this dimension, the model is required to recognize the action shown in the videos. Not only the ability of capture temporal dynamics, but also the knowledge of physical motions, human actions and dynamic interaction between objects is evaluated. 
    \item Action Prediction. The target of this dimension is to predict the future action through the preceding video segment, which requires the understanding of contextual information from videos and temporal reasoning. 
    \item Procedure Understanding. This dimension requires the model to capture all the key actions and perform temporal ordering on them. We aims to evaluate the ability of temporally fine-grained understanding and procedure reasoning. 
\end{itemize}

\begin{figure}
    \includegraphics[width=1.0\textwidth]{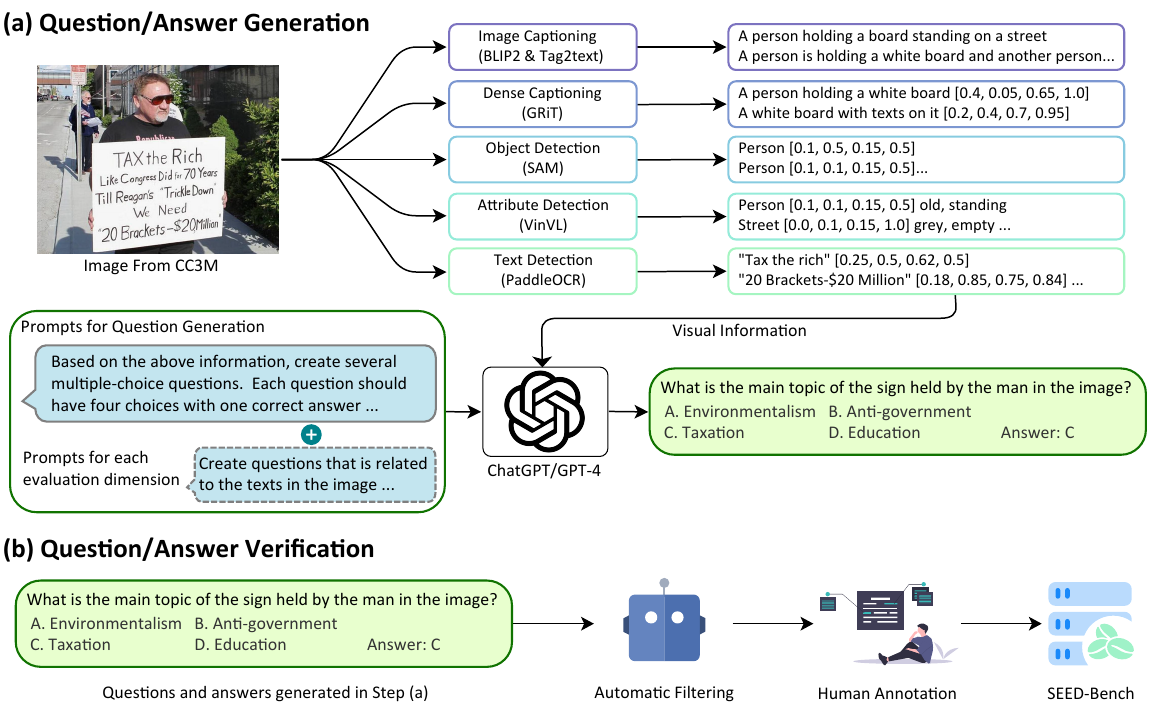}
    \vspace{0pt}
    \caption{Overview of SEED-Bench pipeline for generating multiple-choice questions of images. (a) We first leverage various foundation models to extract visual information including image-level captions, instance-level descriptions and textual elements. Based on specially designed prompts corresponding to specific evaluation dimension, ChatGPT/GPT-4 subsequently  generates questions and four candidate options with one groundtruth answer. (b) We further filter out questions by utilizing LLMs and employ human annotators to select the correct option and classify each question into one evaluation dimension.
    }~\label{fig_pipeline}
        \vspace{-20pt}
\end{figure}

\subsection{Data Source}
\label{sec:source}
To create a benchmark with various evaluation dimensions, we need to collect data containing images with abundant visual information and videos with rich temporal dynamics, so that we can construct diverse challenging multiple-choice questions. In SEED-Bench, we use CC3M~\cite{sharma2018conceptual_gcc} dataset with filtered samples to build questions for spatial understanding. Specifically, considering the noisy original captions of CC3M, we generate captions for each image with Tag2Text~\cite{huang2023tag2text}. We filter out those images with no more than 5 nouns in their captions, so as to ensure the information richness in the remaining images for constructing questions.

We further adopt Something-Something-v2 (SSV2)~\cite{ssv2}, Epic-kitchen 100~\cite{epickitchen100} and Breakfast~\cite{breakfast} dataset to build questions for temporal understanding. SSV2 is an action recognition dataset including 174 fine-grained categories of basic actions with everyday objects and we adopt 1740 videos from its validation set. We also select 138 long videos from Epic-kitchen 100 dataset with temporally annotated action labels. Moreover, videos and fine-grained action segmentation annotations in Breakfast dataset~\cite{breakfast} are utilized for the procedure understanding task.

\subsection{Multiple-Choice Questions} 
\label{sec:mcq}
As shown in Fig.~\ref{fig_pipeline}, our pipeline for generating multiple-choice questions involves question/answer generation and verification. For generating question/answer pairs, we first leverage various foundation models to extract visual information including image-level captions, instance-level descriptions and textual elements. Based on specially designed prompts corresponding to specific evaluation dimension, ChatGPT/GPT-4 subsequently generates questions and four candidate options with one groundtruth answer. For verifying question/answer pairs, we filter out questions that can be answered correctly by multiple LLMs without resorting to visual information. We further employ human annotators to select the correct option and classify each question into one evaluation dimension. 

\noindent\textbf{Visual Information Extraction.}
For constructing questions related to spatial understanding, we interpret the rich information in each image with texts using multiple pretrained models, so that ChatGPT/GPT-4 can understand the image and create questions accordingly. For constructing questions related to temporal understanding, considering that extracting reliable temporal information from videos (especially fine-grained actions and long-term temporal context) is extremely difficult given existing foundation models, we utilize the ground-truth annotations of video datasets. We will explore how to generate questions based on automatically extracted video information in the future. The extraction of visual information for images includes the following parts:

\begin{itemize}
\item \noindent\textbf{Image Captions.} 
Image captions contain the overall description of an image. 
We employ BLIP2~\cite{li2022blip} and Tag2Text~\cite{huang2023tag2text} to create captions for each image.
The former creates captions for the whole image while the latter generates captions based on descriptions of each instance.
The two models complement each other to depict the image content within a single sentence.

\item \noindent\textbf{Instance Descriptions.}
Besides captions which may ignore specific details in the image, we also extract visual information from images using instance-level descriptions, including object detection, attribute detection, and dense captions.
Specifically, we use SAM~\cite{kirillov2023sam} to segment each instance in the image and obtain their bounding boxes according to the segmentation results.
The object labels are obtained using Tag2Text~\cite{huang2023tag2text}.
Besides, we also utilize attribute detector~\cite{zhang2021vinvl} to obtain the attributes of each instance in the image.
Finally, we employ GRiT~\cite{wu2022grit} to generate dense captions, which describe each detected instance in the image with a short sentence.
These instance-level descriptions are complementary to the image captions, further enriching the visual information of each image. 

\item \noindent\textbf{Textual Elements.}
Besides objects, the texts in the image also contain important information describing the image. 
We employ PaddleOCR~\cite{paddleocr} for detecting textual elements.
\end{itemize}

\begin{figure}
    \includegraphics[width=1.0\textwidth]{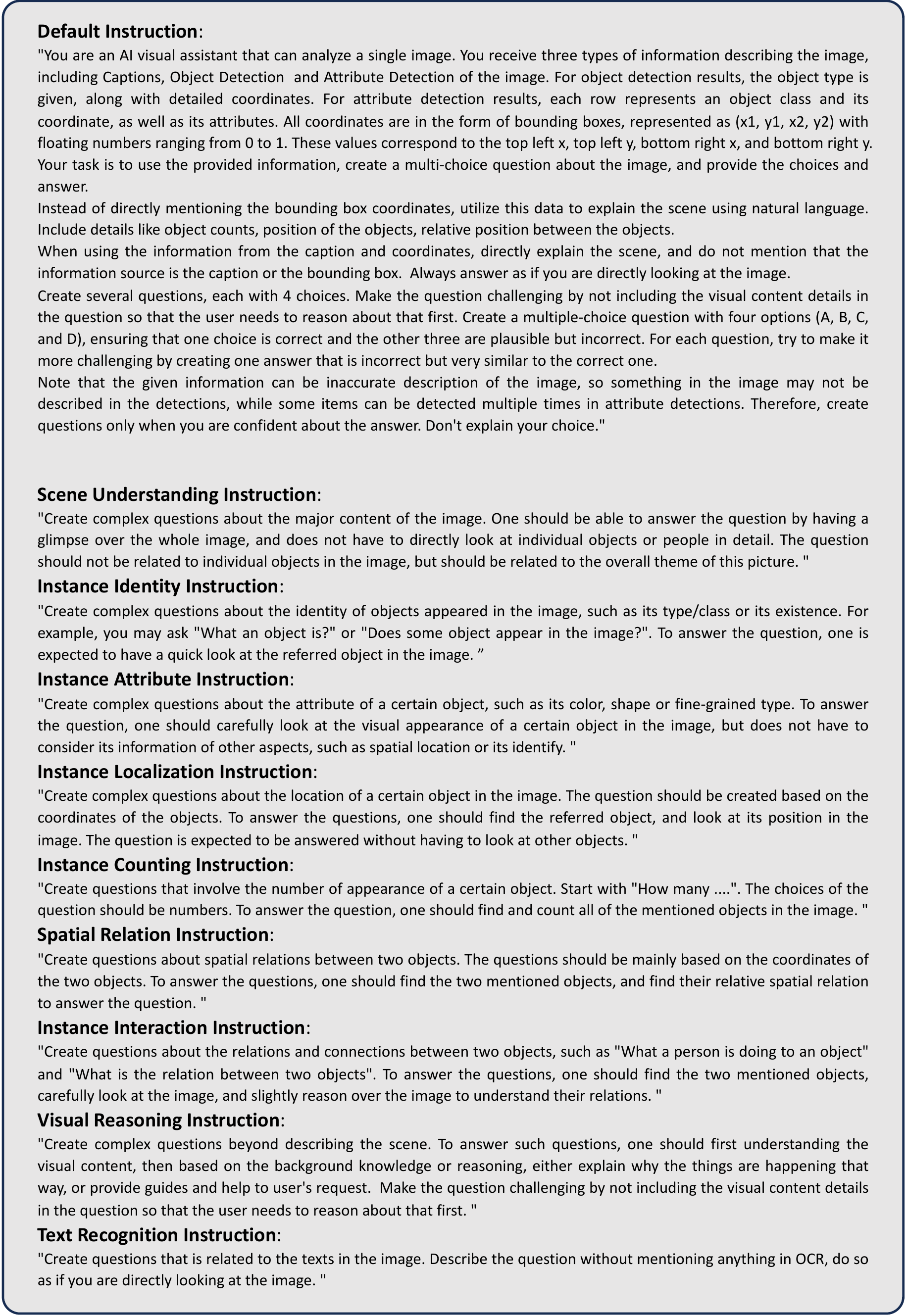}
    \caption{Prompts of generating multiple-choice questions for different evaluation dimensions.}
    \label{fig:prompt}
\end{figure}

\noindent\textbf{Question-Answer Generation.} After extracting visual information from the image and video, we task ChatGPT/GPT-4 with generating multiple-choice questions based on the extracted information or video annotations.  
For each of the spatial understanding evaluation, we carefully design prompts and ask ChatGPT/GPT-4 to create multiple choice questions with four candidate options based on the extracted visual information.
We create questions with ChatGPT for all evaluation dimensions, except for the reasoning dimension, where we use GPT-4~\cite{openai2023gpt4} due to its exceptional reasoning capability. 
For each question, we ask ChatGPT/GPT-4 to create four choices with one correct option and three distractors. We try to make the multiple-choice questions challenging by encouraging the three wrong choices to be similar to the correct one. The detailed prompts of generating multiple-choice questions for different evaluation dimensions are listed in Fig.~\ref{fig:prompt}.
For generating questions related to temporal understanding, we utilize the ground-truth annotations of selected videos as the answer of multi-choice questions and employ ChatGPT to generate three distractors.

\noindent\textbf{Automatic Filtering.}
Our benchmark aims at evaluating the multimodal vision-language understanding capability of MLLMs. 
However, we observe that some generated questions can be correctly answered by LLMs without seeing the image.
We argue that such questions are not helpful to evaluate the visual comprehension capability of MLLMs.
To this end, we feed the generated questions (without image) into three powerful LLMs, including Vicuna-7B~\cite{vicuna}, Flan-T5-XXL~\cite{chung2022scaling_flant5} and LLaMA-7B~\cite{touvron2023llama} and ask them to answer the questions.
We empirically found that $5.52\%$ of the generated questions can be correctly answered by all of the three LLMs.
We filter out these questions from our benchmark.

\noindent\textbf{Human Annotation.}
To ensure the accuracy and objectiveness of SEED-Bench, we further employ human annotators to verify the generated question/answer pairs.  Human annotators are asked to choose the correct answer for each multiple-choice question and categorize each question into one of the evaluation dimension. If one question can not be answered based on the visual input or does not have any correct choice or has multiple correct choices, it will be discarded by human annotators. This results in a clean, high-quality and well-categorized benchmark for evaluation with a total of 19K multiple-choice questions. The statistics of the number of multiple-choice questions in each evaluation dimension is shown in Fig.~\ref{fig:seed_bench_overview}. We can observe a minimum number of questions in text recognition with 85 samples, and a maximum number in instance localization with 4649 samples. We will maintain an even distribution among multiple-choice questions associated with different evaluation dimensions in the future.

\subsection{Evaluation Strategy}
\label{sec:strategy}
Different from MMBench~\cite{liu2023mmbench} that employs ChatGPT to match a model’s prediction to one of the choices in a multiple-choice question (achieves only 87.0\% alignment rate), we adopt the answer ranking strategy~\cite{dai2023instructblip, brown2020gpt3, lin2021truthfulqa} for evaluating existing MLLMs with multiple-choice questions. Specifically, for each choice of a question, we compute the likelihood that an MLLM generates the content of this choice given the question. We select the choice with the highest likelihood as model's prediction. Our evaluation strategy does not rely on the instruction-following capabilities of models to output ``A'' or ``B'' or ``C'' or ``D''. Furthermore, this evaluation strategy eliminates the impact of the order of multiple-choice options on the model's performance.

\section{Evaluation Results}
\begin{table}[]
    \centering
    \caption{Evaluation results of different models on SEED-Bench, where ``Spatial'' shows the averaged performance on nine dimensions for evaluating spatial understanding, and ``Temporal'' shows the averaged performance on three dimensions for evaluating temporal understanding.}\label{tab:performance}
    \vspace{3pt}
    {\small
    \resizebox{\textwidth}{!}{
    \begin{tabular}{cccccccccc}
         \toprule
         \multirow{2}{*}{Model Type} & \multirow{2}{*}{Model} & \multirow{2}{*}{Language Model}& \multicolumn{2}{c}{Spatial} & \multicolumn{2}{c}{Temporal} & \multicolumn{2}{c}{Overall} \\
         \cmidrule(lr){4-9}
         \cmidrule(lr){4-5}
         \cmidrule(lr){6-7}
         \cmidrule(lr){8-9}
         &&& Acc & Rank & Acc & Rank & Acc & Rank \\
         \midrule
         \multirow{3}{*}{LLM} & Flan-T5~\cite{chung2022scaling_flant5} &Flan-T5-XL  &27.32 &17 &28.56 &11 & 27.65 &17 \\
         & Vicuna~\cite{vicuna} &Vicuna-7B &28.16 &16 &29.46 & 8 & 28.50 & 16\\
         & LLaMA~\cite{touvron2023llama} &LLaMA-7B &26.56 &18 &27.27 & 13 &26.75 & 18\\
         \midrule
         \multirow{12}{*}{ImageLLM} & BLIP-2~\cite{li2023blip2} &Flan-T5-XL &49.74 &3 &36.71 & 3 &46.35 &3 \\
         & InstructBLIP~\cite{dai2023instructblip} &Flan-T5-XL &57.80 &2 &\bf38.31 & 1 &52.73 &2\\
         & InstructBLIP Vicuna~\cite{dai2023instructblip} &Vicuna-7B &\bf58.76 &1 &38.05 &2 &\bf53.37 &1\\
         & LLaVA~\cite{liu2023visual_llava} &LLaMA-7B &36.96 &8 &23.75 &16 &33.52 &9\\
         & MiniGPT-4~\cite{zhu2023minigpt4} &Flan-T5-XL &47.40 &4 &29.89 &7 &42.84 &4\\
         & VPGTrans~\cite{2023vpgtrans} &LLaMA-7B &41.81 &5 &31.40 &5 &39.10 &5\\
         & MultiModal-GPT~\cite{gong2023multimodalgpt} &LLaMA-7B &34.54 &12 &29.21 &10 &33.15 &11\\
         & Otter~\cite{li2023otter} &LLaMA-7B &35.16 &11 &30.35 &6 &33.91 &8\\
         & OpenFlamingo~\cite{openflamingo} &LLaMA-7B &34.51 &13 &29.25 &9 &33.14 &12\\
         & LLaMA-Adapter V2~\cite{gao2023llamaadapterv2} &LLaMA-7B &35.19 &10 &25.75 &14 &32.73 &13\\
         & GVT~\cite{wang2023gvt} &Vicuna-7B &35.49 &9 &27.77 &12 &33.48 &10\\
         & mPLUG-Owl~\cite{ye2023mplugowl} &LLaMA-7B &37.88 &7 &23.02 &18 &34.01 &7\\
         \midrule
         \multirow{3}{*}{VideoLLM} & VideoChat~\cite{li2023videochat} &Vicuna-7B &39.02 &6 &33.68 &4 &37.63 &6\\
         & Video-ChatGPT~\cite{maaz2023videochatgpt} &LLaMA-7B &33.88 &14 &23.46 &17 &31.17 &14\\
         & Valley~\cite{luo2023valley} &LLaMA-13B &32.04 &15 &25.41 &15 &30.32 &15\\
         \bottomrule
    \end{tabular}
    }
   }
\end{table}

\subsection{Models}
Based on our SEED-Bench, we evaluate 18 models including 3 LLMs, \textit{i.e.}, Flan-T5~\cite{chung2022scaling_flant5}, Vicuna~\cite{vicuna}, LLaMA~\cite{touvron2023llama}, 12 ImageLLMs, \textit{i.e.}, OpenFlamingo~\cite{openflamingo}, BLIP-2~\cite{li2023blip2}, MiniGPT-4~\cite{zhu2023minigpt4}, LLaVa~\cite{liu2023visual_llava}, mPLUG-Owl~\cite{ye2023mplugowl}, InstructBLIP~\cite{dai2023instructblip}, Otter~\cite{li2023otter}, MultimodalGPT~\cite{gong2023multimodalgpt}, GVT~\cite{wang2023gvt}, PandaGPT~\cite{su2023pandagpt}, VPGTrans~\cite{2023vpgtrans}, LLaMA-Adapter V2~\cite{gao2023llamaadapterv2}, and 3 VideoLLMs, \textit{i.e.}, VideoChat~\cite{li2023videochat}, Video-ChatGPT~\cite{maaz2023videochatgpt} and Valley~\cite{luo2023valley}. Each model is evaluated with all the 12 dimensions including both the spatial and temporal understanding. For ImageLLMs, besides the evaluation of spatial understanding, we aim to investigate their capability to perform temporal reasoning among multiple frames. For VideoLLMs, we seek to explore whether their spatial understanding abilities have degraded by taking a single image as the input.

\begin{figure}
    \includegraphics[width=1.0\textwidth]{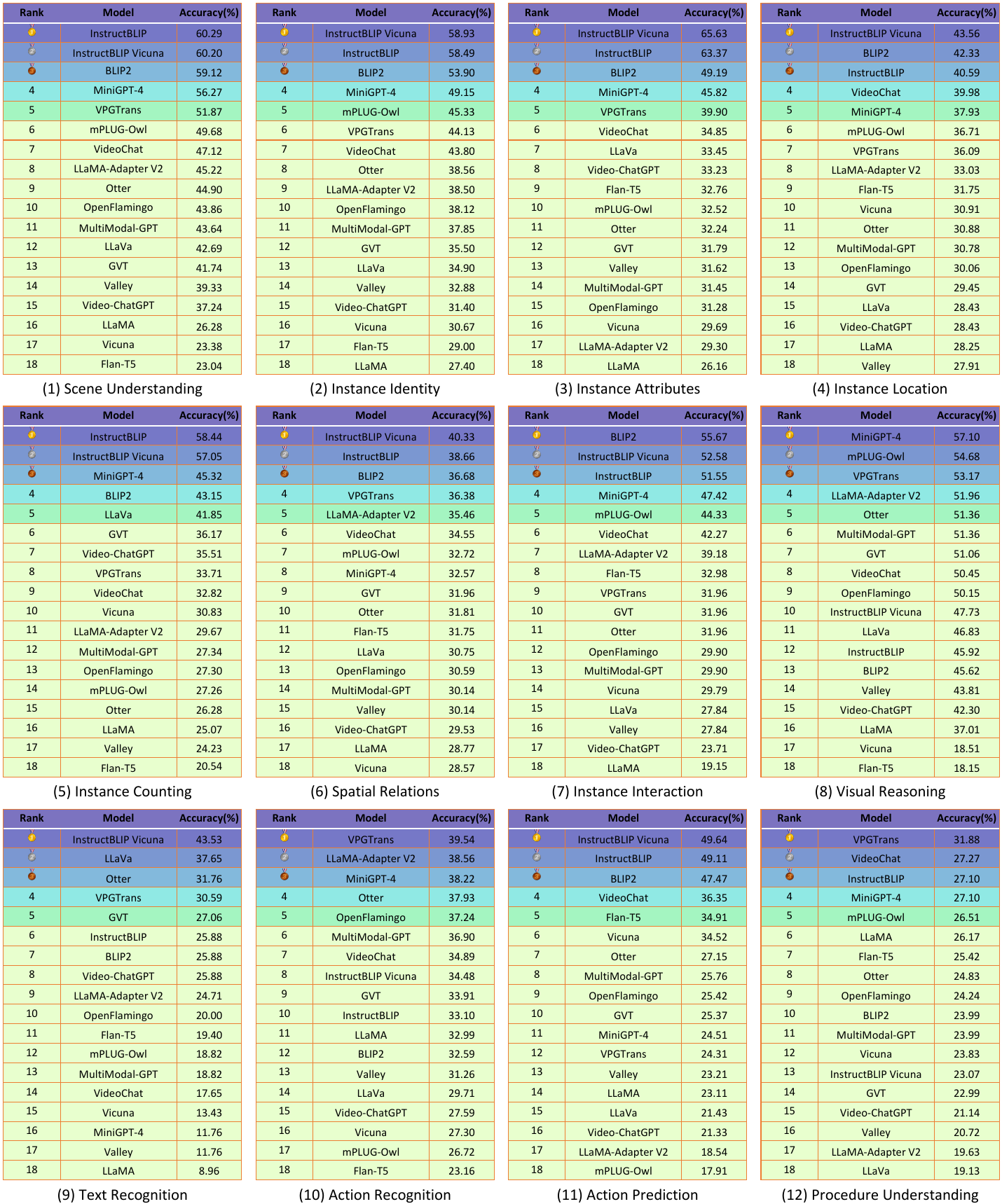}
    \caption{Leaderboards of different evaluation dimensions on SEED-Bench.}
    \label{fig:each_task}
\end{figure}

\subsection{Results}
\begin{figure}
    \includegraphics[width=1.0\textwidth]{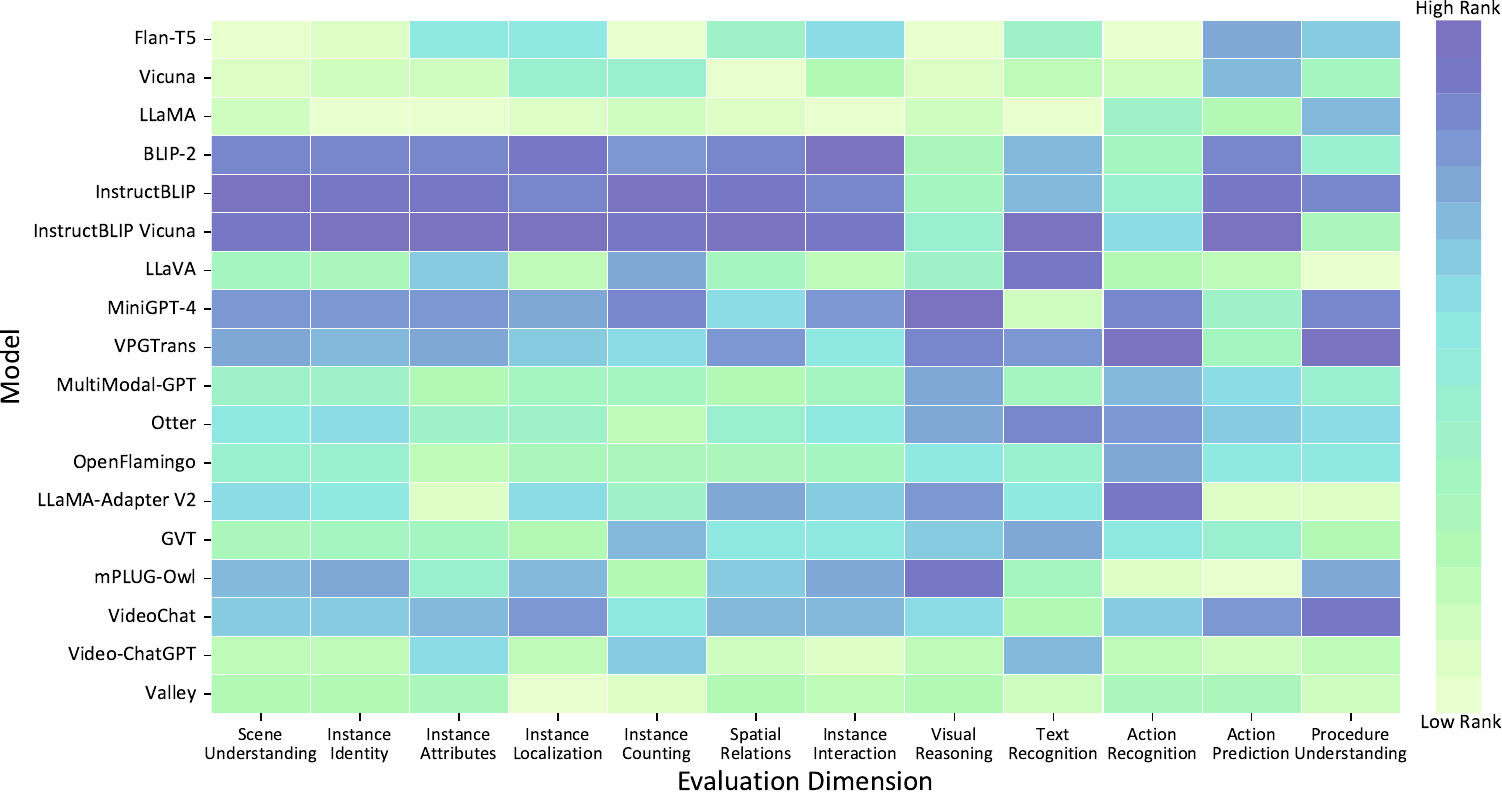}
    \caption{Illustration of each model's performance across different evaluation dimensions, where darker colors represent higher ranks.}
    \label{fig:rank}
    \vspace{-10pt}
\end{figure}

The evaluation results of different models on SEED-Bench are listed in Table.~\ref{tab:benchmark_compare}, where the accuracy refers to the proportion of correctly answered multiple-choice questions relative to the total number of questions. We are surprised to observe that InstructBLIP~\cite{dai2023instructblip} not only achieves the best performance based on the averaged results across nine dimensions for evaluating spatial understanding, but also surpasses VideoLLMs in terms of the averaged results across three dimensions for evaluating temporal understanding. We display leaderboards of various evaluation dimensions on SEED-Bench in Fig.~\ref{fig:each_task} to provide a comprehensive assessment of different models. The overall leaderboard based on the averaged results across all the evaluation dimensions are shown in Fig.~\ref{fig:seed_bench_overview}. To better showcase the the capabilities of models across different evaluation dimensions, we further visualize the ranking of each model within each evaluation dimension in Fig.~\ref{fig:rank}, where darker colors represent higher ranks. We can observe that the BLIP series~\cite{li2023blip2, dai2023instructblip} model achieves competitive results in multiple evaluation dimensions, but they are not good at visual reasoning and action recognition. VideoLLM Valley~\cite{luo2023valley} achieves suboptimal performance in the majority of evaluation dimensions. LLaVa~\cite{liu2023visual_llava} exhibits unparalleled capabilities in the evaluation of text recognition compared to other evaluation dimensions. In terms of specific evaluation dimension, MiniGPT-4~\cite{zhu2023minigpt4} model and mPLUG-Owl~\cite{ye2023mplugowl} model performs better in visual reasoning, while VPGTrans~\cite{2023vpgtrans} model excels in action recognition and procedure understanding. LLaMA Adapter V2~\cite{gao2023llamaadapterv2} model shows more proficiency in action recognition. What's more, Multimodal GPT~\cite{gong2023multimodalgpt}, Otter~\cite{li2023otter}, Openflamingo~\cite{openflamingo}, GVT~\cite{wang2023gvt}, and the three VideoLLMs~\cite{li2023videochat,maaz2023videochatgpt,luo2023valley} exhibit balanced strength across various evaluation dimensions.

\subsection{Analysis}
Through the comprehension and objective evaluation of various models on SEED-Bench, we have observed  a number of findings that can bring insights for future work.

\noindent\textbf{Most MLLMs still exhibit limited performance across all 12 evaluation dimensions.} As shown in Fig.~\ref{fig:seed_bench_overview}, ~\ref{fig:each_task}, most MLLMs (except BLIP series models) can not reach 50\% accuracy on both average performance and the performance on more than three single evaluation dimension. In some specific evaluation dimension (\textit{e.g.}, visual reasoning), it seems that most MLLMs achieve high accuracy. However, when comparing the performance of MLLMs to LLMs, we observe that the performance improvement of most MLLMs is still relatively limited. 

\noindent\textbf{MLLMs achieve relatively high performance on global image comprehension} On the evaluation of scene understanding and visual reasoning, the accuracy of most MLLMs is higher than 40\%, and all MLLMs outperforms LLMs. This shows that MLLMs are more proficient in global understanding and reasoning of images, compared with other evaluation dimensions that require fine-grained instance-level comprehension.

\noindent\textbf{InstructBLIP achieves top performance on 8 of 12 evaluation dimensions. } We can observe that InstructBLIP outperforms other models on 8 evaluation dimensions and the possible explanations for this superior performance are as follows. (a) The instruction-tuning data of InstructBLIP contains totally 16M samples (larger than other instruction-tuning datasets), and covers a wide range of multimodal tasks, even including QA data of OCR and temporal visual reasoning. (b) The weights of LLMs are frozen when performing instruction-tuning of InstructBLIP, which may alleviate catastrophic forgetting. However, InstructBLIP series models still perform poorly on action recognition and procedure understanding that differ significantly from the instruction-tuning data. For instance, on action recognition that requires the understanding of fine-grained actions in Something-Something-v2, InstructBLIP series models can not achieve significant performance gain compared to LLMs (\textit{i.e.}, lower than 2\%). This indicates that InstructBLIP series models may fail to generalize well on the out-of-distribution data. 

\noindent\textbf{MLLMs show weaker abilities in understanding spatial relationships between objects.} The top-ranked model InstructBLIP only achieves 40\% accuracy on the evaluation of spatial relations, which shows that recognizing relative spatial relationships between instances is challenging because there can be many possible arrangements and combinations of spatial relationships between instances. Additionally, spatial relationships between objects may cause ambiguity in some cases, making it difficult to determine their relationship.

\noindent\textbf{Most MLLMs show poor performance for text recognition.} Apart from InstructBLIP, all other models achieve an accuracy lower than 40\% for text recognition due to the lack of textual elements in multimodal pre-training datasets. Since the ability to accurately identify and extract text from images is important, future work should develop models that are better equipped to handle text recognition by pre-training on datasets with rich textual elements in visual data.

\noindent\textbf{VideoLLMs achieve promising results on spatial understanding.} For example, VideoChat achieves 39.98\% accuracy (ranking 4-th on instance localization, surpassing LLaVa by 11.55\% and performing only 3.58\% lower than the top-1 model. It shows that VideoChat's ability of spatial understanding does not degrade by jointly training on both image and video data during the pre-training and instruction-tuning stages.

\noindent\textbf{Most MLLMs exhibit unsatisfactory performance on fine-grained temporal understanding.} It is notable that on the evaluation of procedure understanding, the top-ranked model, VPGTrans, achieves an accuracy that is only 5\% higher than that of LLaMA. The performance improvement of the following 4 MLLMs is even less than 1.2\% compared with LLaMA. This demonstrates that it is extremely difficult for both the ImageLLMs and VideoLLMs to perform fine-grained temporal reasoning so that they can recognize and sort the key actions in a video.

\noindent\textbf{VideoLLMs fail to achieve competitive performance on temporal understanding.}  Although VideoLLMs are instruction-tuned on video data, they do not exhibit a significant advantage on evaluation dimensions for temporal understanding.  Surprisingly, two VideoLLMS (Video-ChatGPT and Valley) even perform worse than most ImageLLMs on action recognition, action prediction and procedure understanding. It indicates that the capabilities of existing VideoLLMs for fine-grained action recognition, temporal relationship understanding and temporal reasoning are still limited. Similar concerns about existing VideoLLMs are also presented in recent works~\cite{li2023videochat,maaz2023videochatgpt}.

\section{Conclusion}
In this work, we propose a large-scale benchmark SEED-Bench to provide a comprehensive and objective evaluation of Multimodal Large Language Models (MLLMs) on generative comprehension. SEED-Bench consists of 19K multiple-choice questions with accurate human annotations, which covers 12 evaluation dimensions for both the spatial and temporal understanding. We design an advanced pipeline to create multiple-choice questions that target specific evaluation dimensions, facilitating the scalability of evaluation data across a variety of domains. We also integrate automatic filtering and manual verification to improve the quality of the generated questions and answers. We conduct a thorough evaluation of 18 models, analyzing and comparing their performances to provide insights for future research. We plan to launch and consistently maintain a leaderboard, offering a platform for the community to assess model performance. We will continue to further broadening the evaluation dimensions of SEED-Bench with more data.

\subsection*{Acknowledgements}
We sincerely acknowledge Junting Pan (CUHK MMLab) for the insightful suggestions, Zhan Tong (Nanjing University) for the data processing, and Yi Chen (Tencent AI Lab) for the engaging discussions.

{\small
\bibliographystyle{unsrt}
\bibliography{egbib}
}

\end{document}